\pdfoutput=1
\documentclass[11pt]{article}

\usepackage[final]{acl}
\usepackage{hyperref}
\usepackage{url}

\usepackage{times}
\usepackage{latexsym}
\usepackage[T1]{fontenc}
\usepackage[utf8]{inputenc}
\usepackage{microtype}
\usepackage{inconsolata}
\usepackage{graphicx}
\usepackage{multirow}
\usepackage{booktabs}
\usepackage{diagbox}
\usepackage{natbib}
\usepackage{algorithm}
\usepackage{algorithmic}
\usepackage{amsmath}
\usepackage{CJK}
\usepackage{wrapfig}
\usepackage{xspace}

\usepackage[most]{tcolorbox}
\usepackage{float}
\usepackage{xspace}
\usepackage{colortbl}
\usepackage{xcolor}
\definecolor{lightmeat}{RGB}{255, 248, 231}
\definecolor{mygray}{gray}{.9}
\definecolor{lightblue}{RGB}{198, 226, 255}
\definecolor{lightred}{RGB}{253, 230, 224}
\definecolor{lightgreen}{RGB}{204, 232, 207}
\definecolor{lightgray}{gray}{0.95}
\definecolor{mediumgray}{gray}{0.85}
\definecolor{darkgray}{gray}{0.7}
\definecolor{cellgreen}{HTML}{0aa858}
\definecolor{cellyellow}{HTML}{ffbc32}
\definecolor{cellred}{HTML}{f4433c}

\tcbset{
aibox/.style={
width=215pt,
top=8pt,
colback=lightmeat,
colframe=black,
colbacktitle=black,
enhanced,
center,
attach boxed title to top left={yshift=-0.12in,xshift=0.15in},
boxed title style={boxrule=0pt,colframe=white},
}
}
\newtcolorbox{AIbox}[2][]{aibox,title=#2,#1}
\newcommand{\method}{\textsc{MM-Detect}\xspace}

\definecolor{pink}{RGB}{255, 20, 147}

\title{\textit{Both Text and Images Leaked!} A Systematic Analysis of Data Contamination in Multimodal LLM}
\author{
    Dingjie Song$^{\dagger,\hspace{.1em}{\textcolor[HTML]{653600}{\boldsymbol{L}}}}$, 
    Sicheng Lai$^{\dagger,\hspace{.1em}{\textcolor[rgb]{0.866,0.639,0}{\boldsymbol{C}}}}$, 
    Mingxuan Wang$^{\hspace{.1em}{\textcolor[rgb]{0.866,0.639,0}{\boldsymbol{C}}}}$, \\
    \textbf{Shunian Chen}$^{\hspace{.1em}{\textcolor[rgb]{0.866,0.639,0}{\boldsymbol{C}}}}$, 
    \textbf{Lichao Sun}$^{\hspace{.1em}{\textcolor[HTML]{653600}{\boldsymbol{L}}}}$$^*$, 
    \textbf{Benyou Wang}$^{\hspace{.1em}{\textcolor[rgb]{0.866,0.639,0}{\boldsymbol{C}}}}$\thanks{Lichao and Benyou are the corresponding authors (\textit{lis221@lehigh.edu,wangbenyou@cuhk.edu.cn}); $^\dagger$ means equal contribution.} \\
    $^{\textcolor[HTML]{653600}{\boldsymbol{L}}}$Lehigh University \\
    $^{\textcolor[rgb]{0.866,0.639,0}{\boldsymbol{C}}}$The Chinese University of Hong Kong, Shenzhen \\
    \textcolor{pink}{\url{https://github.com/MLLM-Data-Contamination/MM-Detect}}
}

\begin{document}
\begin{CJK}{UTF8}{gbsn}
\maketitle

\begin{abstract}

The rapid advancement of multimodal large language models (MLLMs) has significantly enhanced performance across benchmarks. However, data contamination—unintentional memorization of benchmark data during model training—poses critical challenges for fair evaluation. Existing detection methods for unimodal large language models (LLMs) are inadequate for MLLMs due to multimodal data complexity and multi-phase training.
We systematically analyze multimodal data contamination using our analytical framework, \method, which defines two contamination categories—unimodal and cross-modal—and effectively quantifies contamination severity across multiple-choice and caption-based Visual Question Answering tasks. Evaluations on twelve MLLMs and five benchmarks reveal significant contamination, particularly in proprietary models and older benchmarks. Crucially, contamination sometimes originates during unimodal pre-training rather than solely from multimodal fine-tuning. Our insights refine contamination understanding, guiding evaluation practices and improving multimodal model reliability.

\end{abstract}

\section{Introduction}

% Background: Why we need detection for MLLM?
The development of MLLMs has exceeded expectations~\citep{llava-1.5-7b,vila-1.5-3b}, showcasing extraordinary performance on various multimodal benchmarks~\citep{scienceqa,MMBench,song2024milebench}, even surpassing human performance. 
However, due to the partial obscurity associated with MLLMs training~\citep{gpt,gemini,DBLP:conf/icml/Huang0WWZLGHLZL24}, it remains challenging to definitively ascertain the impact of training data on model performance, despite some works showing the employment of the training set of certain datasets~\citep{llava-1.5-7b,internvl,qwen-vl-chat}. 
The issue of data contamination, occurring when training or test data of benchmarks is exposed during the model training phase~\citep{xu2024benchmarking}, could potentially instigate inequitable performance comparisons among models. 
This not only creates a dilemma for users in model selection but also poses a significant hurdle to further advancements in this domain.

% Existing works and their analytical limitations
Existing contamination detection methods only focus on LLMs \citep{DBLP:conf/csfw/YeomGFJ18,ts-guessing,DBLP:conf/acl/DongJLJGYL24}, showing limitations when applied to MLLMs, due to their multimodal data complexity and multi-stage training processes \citep{llava-1.5-7b,li2023blip,DBLP:journals/corr/abs-2503-06072}. Thus, systematic analytical frameworks tailored explicitly for multimodal contamination are urgently needed.

\begin{figure}[t]
    \centering
    \includegraphics[width=1\linewidth]{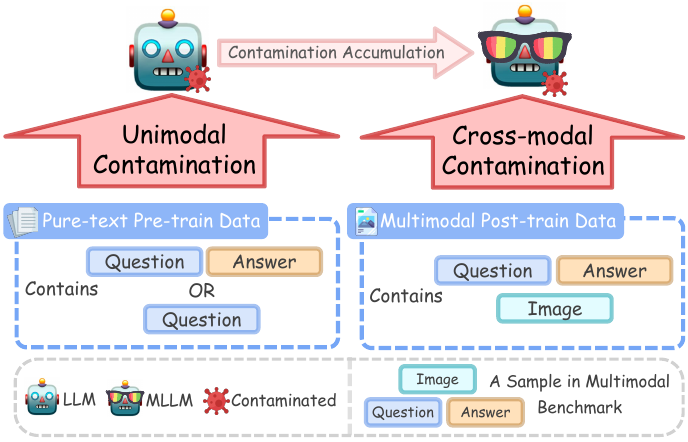}
    % \vspace{-3mm}
    \caption{An analytical breakdown illustrating different forms and origins of multimodal data contamination across distinct training stages of MLLMs.}
    \label{fig:fig1}
    % \vspace{-3mm}
\end{figure}

% Analytical goals and research questions
In this study, we address three key questions:
\begin{itemize}
\item \textbf{How} can we effectively quantify and detect multimodal data contamination?
\item \textbf{What} is the degree of contamination across different MLLMs and benchmark datasets?
\item \textbf{When} is contamination predominantly introduced—during unimodal pre-training or multimodal fine-tuning?
\end{itemize}

% Analytical process
% (1) Definition of Multimodal Data Contamination
To comprehensively answer these questions, we first define \textbf{Multimodal Data Contamination}, as it pertains to the modality of data sources exposed to the MLLMs, into two categories: \textit{Unimodal Contamination} and \textit{Cross-modal Contamination}, as illustrated in Figure~\ref{fig:fig1}.
% (2) Detection framework: Is the multimodal model contaminated?
Subsequently, we unveil a detection framework designed explicitly as an analytical tool, \textbf{\method}, which incorporates two methods, \textit{Option Order Sensitivity Test} and \textit{Slot Guessing for Perturbed Caption}, designed to handle two common types of Visual Question Answering (VQA) tasks: multiple-choice and caption-based questions, respectively. 

% (3) Intentional Contamination
To corroborate the validity and sensitivity of our approach, we deliberately induce contamination in MLLMs, simulating realistic contamination scenarios. Experimental results demonstrate the effectiveness of \method in identifying varying contamination degrees. 
% (4) Wide Experiments
Our evaluations on twelve widely-used MLLMs across five prevalent VQA datasets reveal significant contamination among both proprietary and open-source models. 
% (5) Contamination phase: When was the contamination introduced?
Critically, contamination is not only prevalent in multimodal training data but also can originates from unimodal pre-training phases, impacting older benchmarks disproportionately.

In summary, this work provides the first systematic analytical examination of multimodal data contamination, making the following explicit analytical contributions:
\begin{itemize}
    \item We analytically characterize multimodal contamination into clearly defined unimodal and cross-modal categories, introducing \method as an essential analytical tool.
    \item We systematically quantify how benchmark leakage inflates performance metrics, providing clear insights into dataset and model susceptibility to contamination.
    \item We present novel analytical insights indicating that contamination not solely emerges during the multimodal training stage but could also from unimodal pre-training stage, critically refining current understandings of contamination dynamics.
\end{itemize}

\section{Preliminaries}
\label{sec:prelim}

% \begin{figure*}[t]
%     \centering
%     \vspace{-14mm}
%     \includegraphics[width=1\linewidth]{figs/Figure1.pdf}
%     \vspace{-6mm}
%     \caption{A description of \textbf{Multimodal Data Contamination} (left) and the overview of proposed \textbf{\method} framework (right).}
%     \label{fig:pipeline}
%     \vspace{-4mm}
% \end{figure*}

We formally define the multimodal data contamination and outline the unique challenges associated with its detection.

\subsection{Definition of Multimodal Data Contamination}
\label{sec:definition}

In contrast to single-modal contamination, multimodal contamination may arise from both unimodal and multimodal data sources, as depicted in Figure~\ref{fig:fig1}.
The training data for MLLMs generally consists of pure text pre-training data \( D_{\text{pretrain}} \) and multimodal alignment or instruction-following data \( D_{\text{vision}} \). Consider an instance \( (x, i, y) \) from a benchmark dataset \( D \), where \( x \) represents the text input, \( i \) is the image input, and \( y \) is the label. Data contamination in MLLMs can be categorized into the following two cases:
\begin{itemize}
    \item \textbf{Unimodal Contamination}:  The pair \( (x, y) \) or the input \( x \) appears in \( D_{\text{pretrain}} \).
    \item \textbf{Cross-modal Contamination}:  The triplet \( (x, i, y) \) or the pair \( (x, i) \) appears in \( D_{\text{vision}} \).
\end{itemize}

In both cases, models trained on these data may gain an unfair advantage.

\subsection{Challenges in Multimodal Detection} 
\label{sec:challenges}

The challenges of multimodal contamination detection mainly arise from two aspects.

\paragraph{Challenge I: Inefficiency of Unimodal Methods.} 
Despite the prevalence of unimodal detection methods, their application in multimodal scenarios often encounters difficulties.
% Retrieval-based methods
For example, \textbf{retrieval-based methods}~\cite{DBLP:journals/corr/abs-2005-14165,DBLP:journals/corr/abs-2307-09288} attempt to detect contamination by retrieving large-scale corpora used for model training. Yet, they struggle when retrieving multimodal information.
% Logits-based methods
Similarly, \textbf{logits-based methods}~\citep{DBLP:conf/iclr/ShiAXHLB0Z24,DBLP:conf/csfw/YeomGFJ18} rely on observing the distribution of low-probability tokens in model outputs, but the disparity in token probability distributions is less pronounced in instruction-tuned MLLMs.
% Masking-based methods
\textbf{Masking-based methods}~\cite{ts-guessing}, which assess training contamination by evaluating a model's ability to predict specific missing or masked text, face challenges when images in multimodal samples provide clues, leading to overestimated contamination detection.
% Comparison-based methods
Finally, \textbf{comparison-based methods}~\cite{DBLP:conf/acl/DongJLJGYL24} that measure contamination by comparing model outputs with benchmark data prove to be ineffective for image caption tasks due to low output similarity.
% Our experiments
To validate these inefficiencies, we have conducted comprehensive experiments with compelling results, which are detailed in Appendix~\ref{sec:inefficiency}.

\begin{figure*}[t]
    \centering
    \includegraphics[width=1\linewidth]{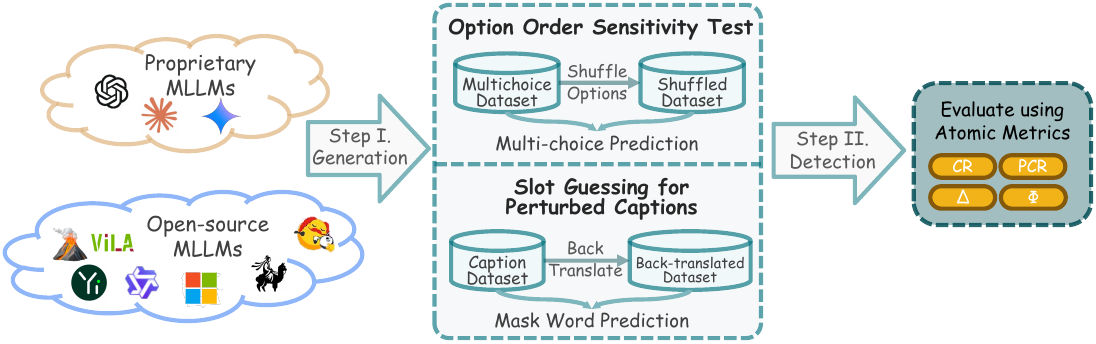}
    \vspace{-2mm}
    \caption{The overview of proposed \textbf{\method} framework.}
    \label{fig:framework}
    \vspace{-2mm}
\end{figure*}

\paragraph{Challenge II: Multi-stage Training in MLLMs.}
Another challenge in detecting contamination in MLLMs is the multi-stage nature of their training~\citep{yin2023survey}. Each stage may be subject to data contamination.
1) Initially, the \textbf{pretraining corpus} could contain the textual components of questions from benchmark samples. Moreover, in certain native multimodal model training~\citep{gemini}, samples may be entirely exposed.
2) Subsequently, during \textbf{multimodal fine-tuning}, the model may utilize training samples of some benchmarks, leading to skewed performance improvements.
3) Furthermore, some models employ extensive mixed image-text data from the internet for \textbf{modality alignment training}~\citep{vila-1.5-3b,qwen-vl-chat}, potentially introducing additional contamination.
Given the challenges, the development of an effective detection framework for multimodal contamination becomes an urgent need.

Based on the discussion above, we have designed a detection method specifically tailored for multimodal contamination, with a particular focus on VQA tasks. Additionally, we have developed a heuristic method to trace the introduction of contamination across different training phases.

\section{Detection Framework: \method}
\label{sec:methods}

We introduce the multimodal contamination detection framework, \textbf{\method}, designed explicitly to support our systematic analysis of contamination phenomena. The core philosophy of \method is to detect the unusual discrepancies in model performance before and after semantic-irrelevant perturbations. 
As depicted in Figure~\ref{fig:framework}, this framework operates in two primary steps: 
\begin{itemize}
    \item The first step is to generate perturbed datasets using two  methods: \textit{Option Order Sensitivity Test} (\S\ref{sec:oost}) and \textit{Slot Guessing for Perturbed Captions} (\S\ref{sec:sgpc}), tailored for multiple-choice and image captioning tasks, respectively.
    
    \item The second step involves the application of predefined metrics to detect contamination (\S\ref{sec:am}), conducting thorough analyses at both the dataset and instance levels.
\end{itemize}

\subsection{Option Order Sensitivity Test}
\label{sec:oost}

\begin{figure}[h]
    \centering
    % \vspace{-2mm}
    \includegraphics[width=1\linewidth]{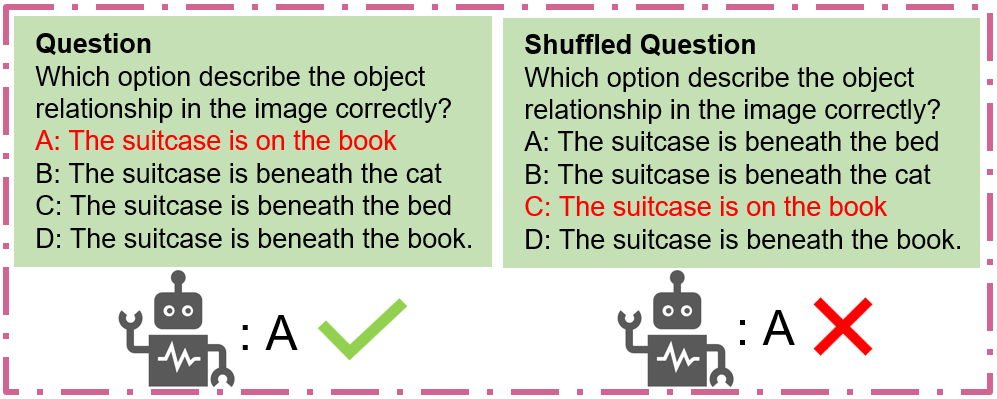}
    % \vspace{-1mm}
    \caption{An example of \textbf{Option Order Sensitivity Test} applied to a contaminated model.}
    % \vspace{-2mm}
    \label{fig:shuffle}
\end{figure}

This method is based on a reasonable and intuitive premise that if the model's performance is highly sensitive to the order of the options, as shown in Figure \ref{fig:shuffle}, it indicates potential contamination, leading the model to memorize a certain canonical order of the options.

\paragraph{Method Formulation.} 
Let \( D \) be a dataset consisting of \( n \) datapoints. Each datapoint \( d_i \) (\( i \in \{1, \ldots, n\} \)) comprises a question \( Q_i \), an associated image \( I_i \), and a set of answer choices \( A_i = \{a_i^1, a_i^2, \ldots, a_i^m\} \), where \( m \) is the number of choices and the correct answer is denoted by \( a_i^c \).

To introduce positional variation, the set \( A_i \) is randomly shuffled to obtain a new set \( A_i' \), ensuring that the index of the correct answer \( a_i^c \) in \( A_i' \) differs from its original position in \( A_i \). The final prompts, before and after shuffling, are constructed by concatenating the image, question and choices:
\[
P = \text{Concat}(I_i, Q_i, A_i),
\]
\[
P' = \text{Concat}(I_i, Q_i, A_i'),
\]
where \( P \) and \( P' \) are the inputs to the model, and \( Q_i \) and \( I_i \) remain unchanged throughout this process.

% \paragraph{Method Formulation.} 
% Let \( D \) represent a dataset comprising \( n \) datapoints. For each datapoint \( d_i \), where \( i \in \{1, \ldots, n\} \), there is a question symbolized by \( Q_i \), an image represented by \( I_i \), and a list of choices is denoted by \( A_i \), such that \( A_i = \{a_i^1, a_i^2, \ldots, a_i^m\} \) and \( m \) is the number of choices for that datapoint. The correct answer is symbolized as \( a_i^c \), where \( a_i^c \in A \).
% The list \( A \) is randomly shuffled to generate \( A' \), ensuring that the index of the correct answer \( a_i^c \) in \( A' \) differs from its index in \( A \), thereby altering the correct answer's position. The final prompts, both before and after the shuffling, are the concatenation of the image, question, and choices:
% \[
% P = \text{Concat}(I, Q, A),P' = \text{Concat}(I, Q, A'),
% \]
% where \( P \) and \( P' \) are the prompts fed into the model, and \( Q \) and \( I \) remain constant.

\subsection{Slot Guessing for Perturbed Caption}
\label{sec:sgpc}

This method is based on the intuition that if a model can predict a missing and important part of a sentence but fails with the back-translated version (from English to Chinese, then back to English), it likely indicates that the model has encountered the original sentence during training.

\begin{figure}[h]
    \centering
    % \vspace{-1mm}
    \includegraphics[width=0.7\linewidth]{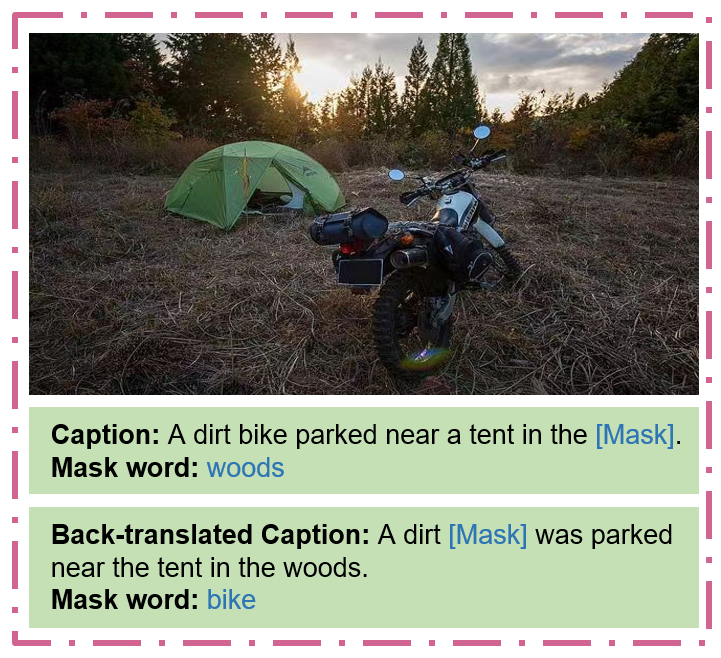}
    % \vspace{-1mm}
    \caption{A simple example shows the procedure.}
    % \vspace{-1mm}
    \label{fig:cs-guessing}
\end{figure}

As shown in Figure \ref{fig:cs-guessing}, the keywords identified are ``woods'' and ``bike''. Since the image contains ``woods'', a correct guess by the model may stem from its multimodal capabilities rather than data contamination. However, if the model fails to predict ``bike'', which is also present in the image, this may indicate potential leakage of this instance.

\paragraph{Method Formulation.} 
Let \( D \) be a dataset containing \( n \) datapoints. Each datapoint \( d_i \) (\( i \in \{1, \ldots, n\} \)) consists of an image-caption pair, where the caption \( S_i \) describes the visual features of the corresponding image \( I_i \). We first apply a back-translation function, where we use the Google Translate API for Python to implement back-translation, to \( S_i \):\footnote{A quantitative analysis of the semantic and lexical similarity between the original and back-translated captions is provided in Appendix~\ref{app:bt_similarity}.}
\[
S'_i = f_{\text{back-translate}}(S_i).
\]
resulting in a paraphrased version \( S'_i \). Next, we perform keyword extraction\footnote{We employ the Stanford POS Tagger \citep{DBLP:conf/emnlp/ToutanvoaM00}, targeting nouns, adjectives, and verbs, as they encapsulate the core meaning of the sentences.} on both \( S_i \) and \( S'_i \):
\[
K_i = f_{\text{keyword}}(S_i), \quad K'_i = f_{\text{keyword}}(S'_i),
\]
where \( K_i \) and \( K'_i \) denote the extracted keywords from \( S_i \) and \( S'_i \), respectively. We then apply a masking function \( f_{\text{mask}} \) to replace the extracted keywords with a placeholder token \texttt{[MASK]}:
\[
S_{i, \text{mask}} = f_{\text{mask}}(S_i, K_i),~S'_{i, \text{mask}} = f_{\text{mask}}(S'_i, K'_i).
\]
The final prompt guiding the model to complete the masked-word prediction can be represented as:
\[
P_i = \text{Concat}(I_i, Q_i, S_{i, \text{mask}}),
\]
% \vspace{-6mm}
\[
P'_i = \text{Concat}(I_i, Q_i, S'_{i, \text{mask}}).
\]

% \paragraph{Method Formulation.} 
% Let \( D \) be a dataset containing \( n \) datapoints. For each datapoint \( d_i \), \( i \in \{1, \ldots, n\} \), there is a corresponding caption \( S_i \) describing the image features. We first apply a back-translation function\footnote{We use Google-Translate API for Python to implement the back-translation.} to \( S_i \):
% \[
% S' = f_{\text{back-translate}}(S)
% \]
% to obtain the back-translated sentence \( S' \). Next, we perform keyword extraction\footnote{We employ the Stanford POS Tagger \citep{DBLP:conf/emnlp/ToutanvoaM00}, targeting nouns, adjectives, or verbs, as they encapsulate the sentences' core meaning.} on both \( S \) and \( S' \):
% \[
% K = f_{\text{keyword}}(S), K' = f_{\text{keyword}}(S'),
% \]
% where \( K \) and \( K' \) are the keywords extracted from \( S \) and \( S' \), respectively. We then use a masking function \( f_{\text{mask}} \) to replace the keywords in the sentences with \texttt{[MASK]}:
% \[
% S_{\text{mask}} = f_{\text{mask}}(S, K), S'_{\text{mask}} = f_{\text{mask}}(S', K').
% \]
% The final prompt can be represented as:
% \[
% P = \text{Concat}(I, Q, S_{\text{mask}}), P' = \text{Concat}(I, Q, S'_{\text{mask}}),
% \]
% where \( I \) is the image and \( Q \) is the instruction guiding the model to complete the mask word prediction task.

% \begin{figure*}
%     \centering
%     \includegraphics[width=1\linewidth]{figs/pipeline.png}
%     \caption{An overview of our \textbf{Contamination Detecting Pipeline}.}
%     \label{fig:pipeline}
% \end{figure*}

\subsection{Detection Metrics}
\label{sec:am}
Detection Metrics serve as the core analytical instruments within \method.
Having introduced two detection methods, we now delineate the atomic metrics for the detection pipeline, which consists of two primary steps.

\paragraph{Step 1: Correct Rate Calculation.} 
This step assesses the model's performance on benchmark \(D\) before and after perturbation. We denote the correct rate (CR) and perturbed correct rate (PCR) uniformly for both Option Order Sensitivity Test (using Accuracy) and Slot Guessing (using Exact Match). Here, \(N\) and \(N'\) are the counts of correct answers before and after perturbation, respectively. They are calculated as:
% \vspace{-2mm}
\[
CR = \frac{N}{|D|}, \quad PCR = \frac{N'}{|D|}.
\]

\paragraph{Step 2: Contamination Degree Analysis.} 
This step quantifies the model's contamination degree based on the performance variation pre- and post-perturbation. Specifically, we introduce two metrics to evaluate contamination at both dataset and instance levels.

\textbf{Dataset Level Metric.}
We evaluate the reduction in atomic metrics, denoted as \( \Delta \):
\[
\Delta = PCR - CR 
\]
This reduction indicates the model's familiarity or memory of the original benchmark relative to the perturbed set, thereby offering insights into potential contamination at the \textbf{dataset level}. A significant negative \( \Delta \) suggests potential extensive leakage in the benchmark dataset, leading to highly perturbation-sensitive model performance.

\textbf{Instance Level Metric.}
Despite a non-significant or positive \( \Delta \), contamination may still occur at the instance level, as some instances may still have been unintentionally included during training. To identify such instances, we compute \( X \), the count of cases where the model provided correct answers before perturbation but incorrect answers after. The \textbf{instance leakage metric \( \Phi \)} is then obtained by dividing \( X \) by the dataset size:
\[
\Phi = \frac{X}{|D|},
\]
where a larger \( \Phi \) indicates a higher likelihood of instance leakage. 
% The entire pipeline, including how these levels are assigned, is summarized in Algorithm~\ref{algo:1}.
% % 
% % Algorithm \ref{algo:1} presents the complete workflow of \method. The rationale for the leakage level and the threshold of \( \Delta \) is based on the experiment in \S\ref{sec:contexperiment}.

% \begin{center}
% \begin{minipage}{0.8\linewidth}
% \vspace{-3mm}
% \begin{algorithm}[H]
%     \footnotesize
%     \caption{\revise{\method Framework}} \label{algo:1}
%     \begin{algorithmic}[1]
%         \REQUIRE Benchmark dataset $D$, Model $M$
%         \IF{$D$ is multiple-choice}
%         \STATE Generate perturbed set $D_{\text{pert}}$ via \S\ref{sec:oost}
%         \ELSE
%         \STATE Generate perturbed set $D_{\text{pert}}$ via \S\ref{sec:sgpc}
%         \ENDIF
%         \STATE Compute $CR$, $PCR$, $\Delta$, $\Phi$ using \S\ref{sec:am}
%         \ENSURE $CR$, $PCR$, $\Delta$, $\Phi$
%     \end{algorithmic}
% \end{algorithm}
% \end{minipage}
% \end{center}

Compared to methods relying solely on accuracy or perplexity, \method explicitly highlights performance drop after perturbations, preventing exaggeration or underestimation of contamination. Moreover, it offers advantages of lower computational overhead, higher sensitivity, and effective black-box applicability, thus serving as an essential analytical toolkit in our study.

\section{Evaluating \method with Intentional Contamination}
\label{sec:contexperiment}

This section tackles our first overarching research question — \textbf{How can we effectively quantify and detect multimodal data contamination?} To operationalise this goal, we break RQ1 into three sub‑questions:

\noindent\textbf{SQ1} (Effectiveness) Is \method able to detect contamination regardless of where it is injected?\\
\textbf{SQ2} (Sensitivity) How finely can \method measure different leakage levels?\\
\textbf{SQ3} (Bias Diagnostic) When training-set data leak, can \method reveal the evaluation bias?

We answer these sub-questions by adopting the LLaVA framework and training a suite of 7B-parameter models with intentionally contaminated data during the visual-instruction tuning phase. The contamination protocol and data split follow \S\ref{sec:exp_setup}.

\subsection{\method is An Effective Detector}
\label{sec:effective}

We reproduced the LLaVA-1.5-7B experiment to obtain a baseline model without contamination. {Recognizing that contamination can occur anywhere in the training data, we inserted contaminated samples into the visual instruction tuning dataset (\(D_{\text{tuning}}\)) at three positions, early, mid, and late}, creating two groups of contaminated training sets using 1340 ScienceQA test samples or 1000 NoCaps validation samples. Corresponding models, termed Early Cont., Mid Cont., and Late Cont., were then trained for comparison with the baseline.

\begin{table}[h]
    \centering
    \resizebox{0.85\linewidth}{!}{
    \begin{tabular}{l|ccc|ccc}
        \toprule
       \multirow{2}{*}{Models}  & \multicolumn{3}{c|}{ScienceQA Test Set} & \multicolumn{3}{c}{NoCaps Val. Set} \\
        \cline{2-7}
         & CR & PCR & \( \Delta \) & CR & PCR & \( \Delta \)  \\
        \midrule
        Baseline & 61.4 & 61.5 & 0.01  & 33.0 & 32.1 & -0.9 \\
        Early Cont. & 71.5 & 68.1 & \textbf{-3.4}  & 37.5 & 32.0 & \textbf{-5.5} \\
        Mid Cont. & 69.4 & 67.3 & \textbf{-2.1}  & 38.5 & 35.1 & \textbf{-3.4} \\
        Late Cont. & 70.2 & 66.9 & \textbf{-3.3}  & 38.7 & 32.6 & \textbf{-6.1} \\
        \bottomrule
    \end{tabular}
    }
    % \vspace{-2mm}
    \caption{{Detection results on contamination using the ScienceQA test set and NoCaps validation set}.}
    % \vspace{-6mm}
    \label{tab:intentional cont}
\end{table}

Table \ref{tab:intentional cont} shows that incorporating contaminated data during training increases both the model's performance and its sensitivity to perturbations. Compared with the baseline, ScienceQA-contaminated models exhibit average increases in CR and PCR of 9.0\% and 5.9\%, while NoCaps-contaminated models show increases of 5.2\% and 1.1\%. Moreover, all contaminated models demonstrate a marked decrease in \(\Delta\), confirming that \method effectively identifies data contamination.

\subsection{\method is Sensitive and Fine-grained}

We evaluated \method's sensitivity by varying leakage levels in the training set. Using the fully contaminated model as our baseline, we trained additional models with moderate and minimal contamination, by inserting reduced amounts (10\% and 50\%) of contaminated data at the late position of the training set, to assess leakage impact.

\begin{figure}[h]
    \centering
    \includegraphics[width=1\linewidth]{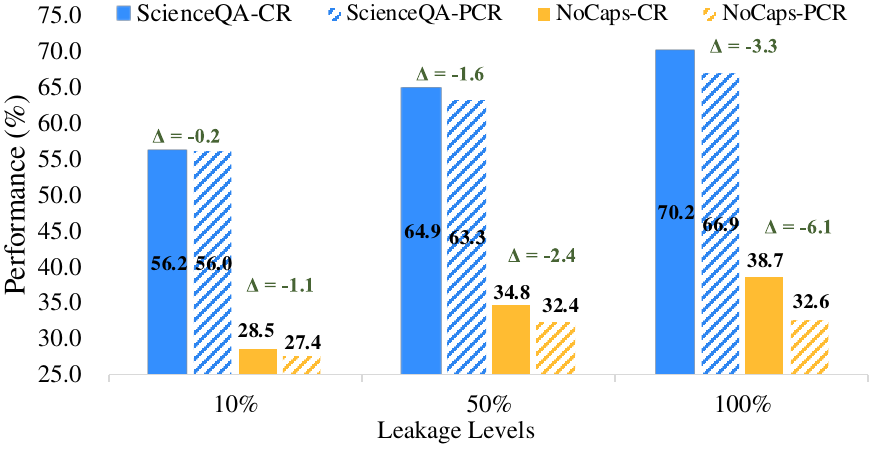}
    % \vspace{-7mm}
    \caption{Performance and atomic metrics evaluated under varying leakage levels on the ScienceQA test set and NoCaps validation set.}
    % \vspace{-3mm}
    \label{fig:scaling law}
\end{figure}

As illustrated in Figure \ref{fig:scaling law}, increasing contamination from 10\% to 50\% to 100\% results in corresponding increases in CR and PCR, alongside progressively larger \(\Delta\) values.
The findings confirm that our framework can accurately differentiate between varying leakage levels in datasets.

% \subsection{Training Set Leakage Leads to Unfairness}
\subsection{\method Diagnoses Evaluation Bias from Training-set Leakage}

We investigated whether \method can detect training set leakage by comparing models trained with and without benchmark data contamination. For the ScienceQA experiment, we appended 2000 ScienceQA training samples to the training dataset, creating a contaminated model. For the COCO experiment, we removed the COCO-Caption2017 training data from the original training dataset, resulting in a model without leakage.

\begin{table}[h]
    \vspace{-1mm}
    \centering
    \small
    \resizebox{0.85\linewidth}{!}{
    \begin{tabular}{l|c|ccc}
        \toprule
        Model & Dataset & CR & PCR & \( \Delta \) \\
        \midrule
        Clean & ScienceQA & 61.4 & 61.5 & 0.01 \\
        Leaked & ScienceQA & 64.3 & 63.8 & \textbf{-0.5} \\
        \midrule
        Clean & COCO-Caption2017 & 32.5 & 31.9 & \textbf{-0.6} \\
        Leaked & COCO-Caption2017 & 38.1 & 34.9 & \textbf{-3.2} \\
        \bottomrule
    \end{tabular}
    }
    \vspace{-2mm}
    \caption{Performance of models trained without (Clean) and with (Leaked) training set contamination.}
    \label{tab:trainset unfairness}
    \vspace{-3mm}
\end{table}

Table \ref{tab:trainset unfairness} compares the models' performance. On the ScienceQA test set, the contaminated model outperforms the clean model by 2.9\% in CR and 2.3\% in PCR, with a $\Delta$ of -0.5. On the COCO-Caption2017 validation set, the model trained with COCO data shows a $\Delta$ of -3.2. The results indicate that training set leakage inflates performance and that \method effectively detects it.

\method also shows strong robustness: $\Delta$ is unaffected by prompt changes (Appendix~\ref{app:prompt-consistency}).
% We further confirm \method’s robustness: $\Delta$ stays effectively unchanged under prompt rewording; see Appendix~\ref{app:prompt-consistency}.

\begin{figure}[h]
\vspace{-1mm}
\begin{AIbox}{Takeaways}\footnotesize
\textit{Both training and test set leakage can result in unfairness, and the degree of contamination can be detected through \method effectively. }
\end{AIbox}
\label{fig:take_away} 
\end{figure}
\vspace{-4mm}

\section{Assessing the Extent of Contamination in MLLMs}
\label{sec:experiment}

In this section, we systematically quantify the extent of contamination across various MLLMs and benchmarks, addressing our second research question — \textbf{What is the degree of contamination?}

\subsection{Setup}
\label{sec:exp_setup}

\paragraph{Models.} 
We conducted evaluations on nine open-source MLLMs, including LLaVA-1.5-7B \cite{llava-1.5-7b}, VILA1.5-3B \cite{vila-1.5-3b}, Qwen-VL-Chat \cite{qwen-vl-chat}, fuyu-8b\footnote{\url{https://www.adept.ai/blog/fuyu-8b}}, idefics2-8b \cite{idefics2-8b}, Phi-3-vision-128k-instruct \cite{phi3}, Yi-VL-6B \cite{yi}, InternVL2-8B \cite{internvl, internvl2-8b}, DeepSeek-VL2-Tiny \cite{deepseek}, as well as three proprietary MLLMs: GPT-4o-2024-08-06 \cite{gpt}, Gemini-1.5-Pro-002 \cite{gemini}, and Claude-3.5-Sonnet-2024-06-20\footnote{\url{https://www.anthropic.com/news/claude-3-5-sonnet}}.
% \lsc{闭源模型测试时间/版本号}

\paragraph{Benchmark Datasets.} 
Our analysis leverages two multi-choice datasets: ScienceQA \cite{scienceqa} and MMStar \cite{mmstar}, along with three caption datasets: COCO-Caption2017 \cite{cococaption}, NoCaps \cite{nocaps}, and Vintage\footnote{\url{https://huggingface.co/datasets/SilentAntagonist/vintage-artworks-60k-captioned}}. MMStar and Vintage, owing to their recent inception, serve to contrast leakage levels with other datasets. We randomly selected 2000 and 1340 samples from ScienceQA's training and test sets, respectively, with 1000 samples from the other datasets. Given the unavailability of public test labels for COCO-Caption2017 and NoCaps, we used their validation sets.

\subsection{Main Results}

\paragraph{Multi-choice Datasets.} 
Table \ref{tab:multi-choice} yields several conclusions:
(1) \textbf{Both open-source and proprietary models exhibit contamination.} For example, on the ScienceQA training set, both open-source models like LLaVA-1.5-7B and idefics2-8b and proprietary Gemini-1.5-Pro show minor contamination degree.
(2) \textbf{Proprietary models are more contaminated.} Claude-3.5-Sonnet, for instance, registers a severe \(\Delta\) with higher \(\Phi\) values on ScienceQA datasets, indicating extensive leakage.
(3) \textbf{Training set leakage is more pronounced than test set leakage.} On the ScienceQA dataset, models generally exhibit larger \(\Delta\) values in the training set, for instance, Claude-3.5-Sonnet shows \(\Delta = -5.3\) on training versus \(-2.4\) on the test set, while most models have near-zero \(\Delta\) on the test set.
(4) \textbf{Older benchmarks are more prone to leak.} The older ScienceQA test set shows more leakage compared to the newer MMStar validation set.

\begin{table*}[t]
    \centering
    \small
    \resizebox{\linewidth}{!}{
    \begin{tabular}{l|rr|rr|rr|rr|rr|rr}
        \toprule
        \textbf{Model} & \multicolumn{4}{c|}{\textbf{ScienceQA Training Set}} & \multicolumn{4}{c|}{\textbf{ScienceQA Test Set}} & \multicolumn{4}{c}{\textbf{MMStar Validation Set}} \\
        \midrule
        Metric & CR & PCR & \( \Delta \) & \( \Phi \) & CR & PCR & \( \Delta \) & \( \Phi \) & CR & PCR & \( \Delta \) & \( \Phi \) \\
        \midrule
        \rowcolor{mygray}\multicolumn{13}{c}{\textit{Open-source MLLMs}}\\
        LLaVA-1.5-7B & 59.7 & 58.6 & \cellcolor{cellgreen!20} -1.1& 12.7 & 60.3 & 61.6 & 1.3& 10.5 & 38.9 & 41.7 & 2.8& 11.0 \\
        VILA1.5-3B & 57.7 & 58.3 & 0.6& 14.5 & 60.3 & 59.8 & \cellcolor{cellgreen!20} -0.5& 14.8 & 38.6 & 37.6 & \cellcolor{cellgreen!20} -1.0& 13.9 \\
        Qwen-VL-Chat & 58.4 & 60.8 & 2.5& 13.3 & 60.3 & 60.4 & 0.1& 13.7 & 40.9 & 44.2 & 3.3& 13.2 \\
        fuyu-8b & 36.5 & 37.5 & 1.0& 13.4 & 37.4 & 36.9 & \cellcolor{cellgreen!20} -0.5& \textbf{14.9} & 28.2 & 27.0 & \cellcolor{cellgreen!20}\textbf{-1.2}& \textbf{17.7} \\
        idefics2-8b & 85.1 & 84.0 & \cellcolor{cellgreen!20} -1.2& 3.7 & 84.0 & 84.3 & 0.3& 2.8 & 48.2 & 49.3 & 1.1& 7.9 \\
        Phi-3-vision-128k-instruct & 90.5 & 90.4 & -0.1& 4.6 & 88.4 & 89.1 & 0.7& 3.9 & 48.7 & 51.9 & 3.2& 7.2 \\
        Yi-VL-6B & 60.5 & 61.8 & 1.3& 10.0 & 59.5 & 61.3 & 1.8& 9.6 & 38.8 & 44.0 & 5.2& 9.3 \\
        InternVL2-8B & 94.1 & 93.9 & \cellcolor{cellgreen!20} -0.3& 2.0 & 92.3 & 93.1 & 0.8& 1.7 & 56.9 & 60.1 & 3.2& 5.1 \\
        DeepSeek-VL2-Tiny & 86.4 & 86.5 & 0.1& 5.3 & 87.1 & 86.9 & \cellcolor{cellgreen!20} -0.2& 5.3 & 51.1 & 52.1 & 1.0& 10.7 \\
        \midrule
        \rowcolor{mygray}\multicolumn{13}{c}{\textit{Proprietary MLLMs}}\\
        GPT-4o & 69.9 & 70.0 & 0.1& 2.7 & 69.1 & 69.7 & 0.6& 2.8 & 48.6 & 50.5 & 1.9& 9.4 \\
        Gemini-1.5-Pro & 68.5 & 67.9 & \cellcolor{cellgreen!20} -0.6& 6.6 & 66.5 & 66.2 & \cellcolor{cellgreen!20} -0.3& 7.1 & 45.7 & 45.5 & \cellcolor{cellgreen!20} -0.2& 9.9 \\
        Claude-3.5-Sonnet & 70.3 & 65.0 & \cellcolor{cellred!20}\textbf{-5.3}& \textbf{15.3} & 67.3 & 64.9 & \cellcolor{cellyellow!20}\textbf{-2.4}& 12.4 & 36.3 & 36.4 & 0.1& 15.9 \\
        \bottomrule
    \end{tabular}
    }
    % \vspace{-3mm}
    \caption[Comparison of MLLMs on multi-choice datasets.]
    {Comparison of MLLMs on multi-choice datasets. Bold values represent the most significant $\Delta$ or $\Phi$; color codes denote contamination degree: {\color{cellgreen}green} for minor leakage, {\color{cellyellow}yellow} for partial leakage, and {\color{cellred}red} for severe leakage.\protect\footnotemark}
    % \vspace{-3mm}
    \label{tab:multi-choice}
\end{table*}
% \footnotetext{Based on intentional contamination experiments in~\S\ref{sec:effective}, the degrees on multi-choice datasets are defined as follows: \(\Delta \in (-1.6, -0.2]\) for minor leakage, \(\Delta \in (-2.9, -1.6]\) for partial leakage, and \(\Delta \leq -2.9\) for severe leakage. Detailed analysis is in Appendix~\ref{app:cont_analysis}}.
\footnotetext{Following \S\ref{sec:effective}: multi-choice leakage levels by $\Delta$—minor ($-1.6<\Delta\le-0.2$), partial ($-2.9<\Delta\le-1.6$), severe ($\Delta\le-2.9$); caption leakage levels by $\Delta$—minor ($-2.4<\Delta\le-1.1$), partial ($-5.0<\Delta\le-2.4$), severe ($\Delta\le-5.0$). See Appendix~\ref{app:cont_analysis}.}

\begin{table*}[t]
    \centering
    \small
    \resizebox{\linewidth}{!}{
    \begin{tabular}{l|rr|rr|rr|rr|rr|rr}
        \toprule
        \textbf{Model} & \multicolumn{4}{c|}{\textbf{COCO Validation Set}} & \multicolumn{4}{c|}{\textbf{NoCaps Validation Set}} & \multicolumn{4}{c}{\textbf{Vintage Training Set}} \\
        \midrule
        Metric & CR & PCR & \( \Delta \) & \( \Phi \) & CR & PCR & \( \Delta \) & \( \Phi \) & CR & PCR & \( \Delta \) & \( \Phi \) \\
        \midrule
        \rowcolor{mygray}\multicolumn{13}{c}{\textit{Open-source MLLMs}}\\
        LLaVA-1.5-7B & 34.6 & 34.0 & -0.6& 19.0 & 30.9 & 28.5 & \cellcolor{cellyellow!20} -2.4& 17.9 & 10.8 & 10.1 & -0.7& 9.0 \\
        VILA1.5-3B & 19.1 & 20.5 & 1.4& 13.0 & 19.1 & 20.5 & 1.4& 13.0 & 1.5 & 2.2 & 0.7& 1.5 \\
        Qwen-VL-Chat & 32.2 & 30.3 & \cellcolor{cellgreen!20} -1.9& 19.2 & 28.7 & 27.3 & \cellcolor{cellgreen!20} -1.4& 16.7 & 15.1 & 15.4 & 0.3& 12.4 \\
        fuyu-8b & 9.6 & 10.6 & 1.0& 7.8 & 10.0 & 9.8 & -0.2& 8.3 & 2.4 & 3.3 & 0.9& 2.3 \\
        idefics2-8b & 43.5 & 42.3 & \cellcolor{cellgreen!20} -1.2& 21.2 & 42.6 & 37.5 & \cellcolor{cellred!20}\textbf{-5.1}& \textbf{23.3} & 18.5 & 17.0 & \cellcolor{cellgreen!20} -1.5& 14.5 \\
        Phi-3-vision-128k-instruct & 38.8 & 39.3 & 0.5& 19.4 & 36.9 & 33.3 & \cellcolor{cellyellow!20} -3.6& 19.7 & 17.4 & 11.7 & \cellcolor{cellred!20}\textbf{-5.7}& 14.3  \\
        Yi-VL-6B & 43.9 & 43.3 & -0.6& 19.4 & 37.2 & 36.1 & \cellcolor{cellgreen!20} -1.1& 17.5 & 3.3 & 4.2 & 0.9& 2.8 \\
        InternVL2-8B & 53.3 & 51.9 & \cellcolor{cellgreen!20} -1.4& 20.4 & 48.0 & 46.2 & \cellcolor{cellgreen!20} -1.8& 20.9 & 28.0 & 28.7 & 0.7& 18.8 \\
        DeepSeek-VL2-Tiny & 23.8 & 21.4 & \cellcolor{cellyellow!20} -2.4& 13.5 & 19.3 & 18.1 & \cellcolor{cellgreen!20} -1.2& 12.2 & 7.5 & 6.9 & -0.6& 6.3 \\
        \midrule
        \rowcolor{mygray}\multicolumn{13}{c}{\textit{Proprietary MLLMs}}\\
        GPT-4o & 58.1 & 54.4 & \cellcolor{cellyellow!20}\textbf{-3.7}& \textbf{23.1} & 54.2 & 55.1 & 0.9& 19.4 & 36.3 & 38.4 & 2.1& 20.1 \\
        Gemini-1.5-Pro & 57.5 & 55.3 & \cellcolor{cellgreen!20} -2.2& 21.6 & 51.2 & 52.0 & 0.8& 18.7 & 46.3 & 41.0 & \cellcolor{cellred!20} -5.3& \textbf{28.3} \\
        Claude-3.5-Sonnet & 53.7 & 51.0 & \cellcolor{cellyellow!20} -2.7& 21.8 & 50.8 & 51.5 & 0.7& 20.0 & 35.2 & 33.0 & \cellcolor{cellgreen!20} -2.2& 21.3 \\
        \bottomrule
    \end{tabular}
    }
    % \vspace{-3mm}
    \caption[Comparison of MLLMs on caption datasets.]
    {Comparison of MLLMs on caption datasets. Bold values represent the most significant $\Delta$ or $\Phi$; color codes denote contamination degree: {\color{cellgreen}green} for minor leakage, {\color{cellyellow}yellow} for partial leakage, and {\color{cellred}red} for severe leakage.}
    % \vspace{-4mm}
    \label{tab:caption}
\end{table*}
% \footnotetext{Based on intentional contamination experiments in~\S\ref{sec:effective}, the degrees on caption datasets are defined as follows: \(\Delta \in (-2.4, -1.1]\) for minor leakage, \(\Delta \in (-5.0, -2.4]\) for partial leakage, and \(\Delta \leq -5.0\) for severe leakage.}

\paragraph{Caption Datasets.} 
Table \ref{tab:caption} yields several conclusions:
(1) \textbf{Both open-source and proprietary models exhibit contamination on caption datasets.} For example, in the COCO Validation Set, open-source models such as DeepSeek-VL2-Tiny and proprietary models like GPT-4o record a significant contamination degree.
(2) \textbf{Leakage levels vary significantly by benchmark.} For example, on the NoCaps Validation Set, open-source models exhibit more pronounced contamination degree than proprietary models, whereas the trend reverses on the COCO Validation Set.
These findings confirm that caption datasets are vulnerable to leakage, with proprietary models generally exhibiting more pronounced contamination effects.

\begin{figure}[h]
% \vspace{-2mm}
\begin{AIbox}{Takeaways}\footnotesize
\textit{Multimodal data contamination, at both dataset and instance levels, is prevalent in open-source and proprietary MLLMs across multi-choice and image caption datasets.}
\end{AIbox} 
% \vspace{-5mm}
\label{fig:take_away_main} 
\end{figure}

% \section{At Which Stage is Contamination Introduced?}
\section{Identifying the Origin of Contamination in MLLMs}
\label{sec:stage}

In this section, we address our third research question — \textbf{When is contamination predominantly introduced?}
Although the training data for some MLLMs is openly documented, an important question remains: if contamination does not arise during the multimodal training phase, could it stem from the unimodal (pre-training) phase, as defined in \S\ref{sec:definition}?
To address this possibility, we examined the underlying LLMs of the evaluated MLLMs and conducted a series of experiments (\S\ref{subsec:heurimethod}). We also explored the origins of cross-modal contamination arising during visual instruction tuning (\S\ref{subsec:vist}).

\subsection{Heuristic Detection of Unimodal Pre-training Contamination}
\label{subsec:heurimethod}

We employed a heuristic approach based on the intuition that if an LLM can correctly answer an \textbf{image-required} question \textbf{without the image} when \textbf{random guessing is effectively inhibited}, it may indicate the leakage of that instance. 

\paragraph{Experiment Setup.}  
We used MMStar as the benchmark, where \textbf{every question relies on visual input for correct answers}. 
The tested models include LLaMA2-7B \cite{llama2} (used by LLaVA-1.5 and VILA), Qwen-7B \cite{qwen} (used by Qwen-VL), Mistral-7B-v0.1 \cite{jiang2023mistral7b} (used by idefics2), Phi-3-small-128k-instruct \cite{phi3} (used by Phi-3-vision), Yi-6B \cite{yi} (used by Yi-VL), and Internlm2-7B \cite{internlm2} (used by InternVL2). 
\textbf{To inhibit random guessing}, we appended the prompt ``\textit{If you do not know the answer, output I don't know}'' to the instructions. {A sanity check in Appendix~\ref{app:sanity-idk} confirms that this uncertainty clause effectively suppresses lucky guesses, validating its inclusion in our main protocol.} Accuracy — the frequency with which models correctly answer questions without image input — is reported as the primary metric. 
Note that we did not evaluate Fuyu-8B and proprietary models since their unimodal LLM components and training data remain undisclosed.

\paragraph{Main Results.}

\begin{table}[h]
    % \vspace{-3mm}
    \centering
    \small
    \resizebox{\linewidth}{!}{
    \begin{tabular}{l|c|c}
        \toprule
        Model & Accuracy  &\color{black}{\( \Phi \)$_{M}$}\\
        \midrule
        LLaMA2-7b (LLaVA-1.5 \& VILA) & 25.6  & \color{black!55}{11.0}
\\
        Qwen-7B (Qwen-VL) & 13.2  &\color{black!65}{13.2}
\\
        Internlm2-7B (InternVL2) & 11.0  &\color{black!15}{5.1}
\\
        Mistral-7B-v0.1 (idefics2) & 10.7  &\color{black!30}{7.9}
\\
        Phi-3-small-128k-instruct (Phi-3-vision) & 6.1  &\color{black!25}{7.2}
\\
        Yi-6B (Yi-VL) & 3.4  &\color{black!40}{9.3}\\
        \bottomrule
    \end{tabular}
    }
    % \vspace{-3mm}
    \caption{Contamination rates of LLMs used by MLLMs. \( \Phi \)$_{M}$ denotes the \( \Phi \) of the respective MLLMs.}
    \label{tab:pretrained}
    % \vspace{-5mm}
\end{table}

Table \ref{tab:pretrained} yields several conclusions:
(1) \textbf{Contamination occurs in LLM.} All models exhibit varied contamination rates, indicating that their pre-training data likely included text from multimodal benchmarks.
(2) \textbf{Elevated LLM contamination correlates with increased MLLM leakage.} For instance, VILA1.5-3B and Qwen-VL-Chat exhibit significant \(\Phi\) values that mirror their underlying LLM contamination levels.
These findings suggest that contamination in these MLLMs may originate partly from the LLMs' pre-training phase, rather than solely from multimodal training.

% \vspace{-3mm}
% \subsection{Tracing Origins: A Review of MLLM's Visual Instruction Tuning Data}
\subsection{Analyzing Cross-modal Contamination in Multimodal Fine-tuning}
\label{subsec:vist}

To investigate the origins of cross-modal contamination, we scrutinize the visual instruction tuning data of MLLMs. We delve into the construction process of three benchmark datasets: ScienceQA, COCO Caption, and Nocaps, comparing them with the training data and its sources of various open-source MLLMs to analyze the degree of overlap.

\begin{table}[h]
    \centering
    \resizebox{1\linewidth}{!}{
    \begin{tabular}{c|ccc}
        \toprule
        Model & ScienceQA & COCO Caption & Nocaps \\
        \midrule
         Phi-3-Vision & \cellcolor{cellgreen!60}0.7& \cellcolor{cellgreen!60}0.5& \cellcolor{cellyellow!60}-3.6\\
         VILA & \cellcolor{cellgreen!60}-0.5& \cellcolor{cellyellow!60}1.4& \cellcolor{cellyellow!60}1.4\\
         Idefics2 & \cellcolor{cellgreen!60}0.3& \cellcolor{cellyellow!60}-1.2& \cellcolor{cellyellow!60}-5.1\\
         LLaVA-1.5 & \cellcolor{cellgreen!60}1.3& \cellcolor{cellyellow!60}-0.6& \cellcolor{cellyellow!60}-2.4\\
         Yi-VL & \cellcolor{cellgreen!60}1.8& \cellcolor{cellyellow!60}-0.6& \cellcolor{cellyellow!60}-1.1\\
         DeepSeek-VL2 & \cellcolor{cellgreen!60}-0.2& \cellcolor{cellyellow!60}-2.4& \cellcolor{cellyellow!60}-1.2\\
         Qwen-VL-Chat & \cellcolor{cellgreen!60}0.1& \cellcolor{cellred!60}-1.9& \cellcolor{cellred!60}-1.4\\
         InternVL2 & \cellcolor{cellred!60}0.8& \cellcolor{cellred!60}-1.4& \cellcolor{cellred!60}-1.8\\
        \bottomrule
    \end{tabular}
    }
    % \vspace{-3mm}
    \caption{Depiction of the overlap between the training data of MLLMs and the benchmarks, as well as the contamination degree \( \Delta \) of MLLMs on benchmarks. {\color{cellgreen} Green} signifies no overlap, {\color{cellyellow} yellow} suggests potential overlap, and {\color{cellred} Red} indicates partial or entire overlap.}
    \label{tab:overlap}
    % \vspace{-3mm}
\end{table}

As Table \ref{tab:overlap} illustrates, MLLMs marked in red and yellow typically exhibit a significant contamination degree. Yet, even MLLMs labeled in green aren't exempt from the risk of cross-modal contamination. This is because some models have been trained on large-scale interleaved image-text datasets (e.g., OBELICS~\citep{OBELICS}), datasets derived from online sources (e.g., Conceptual Caption~\citep{ConceptualCaption}), or in-house data. Furthermore, some models haven't fully disclosed their training data, which may lead to overlooked potential leaks in benchmark datasets.
A detailed overlap analysis is in Appendix~\ref{app:MLLMs Data}.

\begin{figure}[h]
% \vspace{-2mm}
\begin{AIbox}{Takeaways}\footnotesize
\textit{The contamination in MLLMs may not only stem from cross-modal contamination but also from unimodal contamination, both of which can significantly impact the overall performance.}
\end{AIbox} 
% \vspace{-6mm}
\label{fig:take_away_main} 
\end{figure}

\section{Conclusion and Future Work}

In this study, we systematically analyzed multimodal data contamination in MLLMs through our proposed detection framework, \method. We demonstrated that \method effectively quantifies and detects varying contamination degrees, revealing significant performance biases induced by benchmark leakage. Importantly, we identified that contamination originates notably from both unimodal pre-training and multimodal fine-tuning phases, impacting the reliability and fairness of multimodal evaluations.

Future work will focus on two key areas:
\begin{itemize}
    \item Firstly, standardizing the use of multimodal datasets and reporting potential contamination impacts to minimize contamination, thereby enhancing data consistency and quality.
    \item Secondly, creating a continuously updated benchmarking system for the ongoing evaluation of multimodal model performance. 
\end{itemize}
This will support advancements and broader applications in this field.

\section*{Limitations}
We acknowledge several limitations in our work. 
First, this work is limited to discussions around visual modalities, and does not yet cover other modalities such as audio or video.
Second, we only selected widely used and representative multimodal datasets for detection, including multiple-choice datasets and caption datasets, without testing additional datasets, such as open-ended generation and cloze questions. However, we speculate that the method \textit{Slot Guessing for Perturbed Caption} may also apply to other types of image-feature-analyzing benchmarks.
Third, the effectiveness of \textit{Option Order Sensitivity Test} can be undermined by option shuffling, which, while potentially improving model performance, is computationally expensive and may increase the training cost.
Fourth, as a perturbation-based black-box detector, \method might underestimate contamination if a model generalizes sufficiently to answer perturbed questions correctly. Although dataset-level evaluations reduce this risk, completely eliminating such false-negative cases remains an open challenge.

\section*{Acknowledgments}

This work was supported by the Shenzhen Science and Technology Program (JCYJ20220818103001002), Shenzhen Doctoral Startup Funding (RCBS20221008093330065), Tianyuan Fund for Mathematics of National Natural Science Foundation of China (NSFC) (12326608), Shenzhen Science and Technology Program (Shenzhen Key Laboratory Grant No. ZDSYS20230626091302006), and Shenzhen Stability Science Program 2023, Shenzhen Key Lab of Multi-Modal Cognitive Computing.

\bibliography{custom}

\newpage
\appendix
\section{Inefficiency of Unimodal Methods}
\label{sec:inefficiency}
We demonstrate the results of traditional unimodal contamination detection methods applied to MLLMs.

\subsection{Logits-base}
These methods determine contamination by observing the distribution of low-probability tokens in model outputs. However, MLLMs typically undergo instruction fine-tuning, which enhances their instruction-following capabilities, leading to less significant differences in token probability distributions. As shown in Table \ref{tab:logit}, LLaVA-1.5-13b exhibits extremely low perplexity on multimodal benchmark datasets.

\begin{table}[h]
    \centering
    \resizebox{0.8\linewidth}{!}{
    \begin{tabular}{c|c|c}
        \toprule
        Dataset & Perplexity & Split\\
        \midrule
        ScienceQA & 1.4498 & Training Set \\
        MMStar & 1.4359 & Validation Set \\
        COCO-Caption2017 & 1.7530 & Validation Set \\
        NoCaps & 1.8155 & Validation Set \\
        \bottomrule
    \end{tabular}
    }
    \caption{Perplexity of LLaVA-1.5-13b on various multimodal benchmarks (100 samples randomly selected from each dataset).}
    \label{tab:logit}
\end{table}

\subsection{Masking-base}
These methods involve masking phrases or sentences and providing data from the benchmark to guide the model in filling in the missing parts. However, multimodal datasets often contain images that include the masked portions of sentences, effectively providing answers to the model. This results in significantly higher success rates for MLLMs in predicting missing parts compared to unimodal language models, leading to exaggerated contamination detection. As shown in Table \ref{tab:mask}, LLaVA-1.5-13b has a high probability of Exact Match for predicting the masked word.

\begin{table}[h]
    \centering
    \resizebox{1\linewidth}{!}{
    \begin{tabular}{c|c|c|c}
        \toprule
        Dataset & Exact Match & ROUGE-L F1 & Split\\
        \midrule
        COCO-Caption2017 & 0.24 & 0.36 & Validation Set \\
        NoCaps & 0.22 & 0.29 & Validation Set \\
        \bottomrule
    \end{tabular}
    }
    \caption{Contamination detection of LLaVA-1.5-13b using TS-Guessing (question-based) on various multimodal benchmarks (100 samples randomly selected from each dataset).}
    \label{tab:mask}
\end{table}

\subsection{Comparison-base}
These methods identify contamination by comparing the similarity between models' outputs and benchmark data. However, MLLMs often undergo data augmentation, causing their outputs to diverge significantly from the labels in benchmark data, making effective contamination detection challenging. From Table \ref{tab:cdd}, we can see that CDD (Contamination Detection via Output Distribution) indicates a contamination metric of 0\% across all multimodal benchmark datasets.

\begin{table}[h]
    \centering
    \resizebox{1\linewidth}{!}{
    \begin{tabular}{c|c|c}
        \toprule
        Dataset & Contamination Metric & Split\\
        \midrule
        COCO-Caption2017 & 0.0000\% & Validation Set \\
        NoCaps & 0.0000\% & Validation Set \\
        \bottomrule
    \end{tabular}
    }
    \caption{Contamination detection of LLaVA-1.5-13b using CDD (Contamination Detection via Output Distribution) on various multimodal benchmarks (100 samples randomly selected from each dataset).}
    \label{tab:cdd}
\end{table}

\section{Detailed Slot Guessing Pipeline}

\subsection{Back-Translation}

The back-translation function applies a two-step translation process to generate a paraphrased caption \( S'_i \) from the original caption \( S_i \). In this method, we use the Google Translate API to translate the caption into \textbf{Chinese} and then back into the original language to generate the paraphrase.

\begin{algorithm}[H]
\small  
\caption{Back-Translation}
\begin{algorithmic}[1]
\STATE \textbf{Input:} Original caption \( S_i \)
\STATE Translate \( S_i \) to an intermediate language \( L \)
\STATE Translate the resulting caption back from language \( L \) to the original language
\STATE \textbf{Output:} Paraphrased caption \( S'_i \)
\end{algorithmic}
\end{algorithm}

\subsection{Keyword Extraction}

We extract keywords from both the original caption \( S_i \) and the paraphrased caption \( S'_i \) using the Stanford POS Tagger. Keywords are identified as nouns (NN), adjectives (JJ), and verbs (VB), which are considered to encapsulate the core meaning of the sentence. We apply this process to both captions.

\begin{algorithm}[H]
\small  
\caption{Keyword Extraction}
\begin{algorithmic}[1]
\STATE \textbf{Input:} Caption \( S \)
\STATE Apply POS tagging to \( S \) to obtain tags for each word
\STATE Extract words whose POS tags are in \{NN, JJ, VB\}
\STATE \textbf{Output:} List of extracted keywords \( K \)
\end{algorithmic}
\end{algorithm}

\subsection{Keyword Masking}

We apply a masking function to randomly select one keyword from the extracted keywords and replace it with a placeholder token \texttt{[MASK]}. This is done by identifying the position of the selected keyword in the sentence and substituting it with the placeholder.

\begin{algorithm}[H]
\small  
\caption{Keyword Masking}
\begin{algorithmic}[1]
\STATE \textbf{Input:} Caption \( S \), Keywords \( K \)
\STATE \textbf{If} \( K \) is empty \textbf{then} return "failed"
\STATE Randomly select a keyword \( k \) from \( K \)
\STATE Find the first occurrence of \( k \) in \( S \)
\STATE Replace \( k \) with the placeholder \texttt{[MASK]}
\STATE \textbf{Output:} Masked caption \( S_{\text{mask}} \)
\end{algorithmic}
\end{algorithm}

\section{Contamination Degree Analysis}
\label{app:cont_analysis}

Based on ~\S\ref{sec:effective}, the degrees on multi-choice datasets are defined as: \(\Delta \in (-1.6, -0.2]\) for minor leakage, \(\Delta \in (-2.9, -1.6]\) for partial leakage, and \(\Delta \leq -2.9\) for severe leakage. Based on~\S\ref{sec:effective}, the degrees on caption datasets are defined as: \(\Delta \in (-2.4, -1.1]\) for minor leakage, \(\Delta \in (-5.0, -2.4]\) for partial leakage, and \(\Delta \leq -5.0\) for severe leakage. Details are shown in the algorithm~\ref{algo:1}.

\begin{center}
% \begin{minipage}{0.75\linewidth}
% \vspace{-3mm}
\begin{algorithm}[H]
    \footnotesize
    \caption{Contamination Degree Analysis} \label{algo:1}
    \begin{algorithmic}[1]
        \REQUIRE Benchmark dataset $D$, Model $M$
        \STATE Define contamination degree $\mathcal{C}_{\text{Minor}}$, $\mathcal{C}_{\text{Partial}}$, $\mathcal{C}_{\text{Severe}}$
        \IF{$D$ is multiple-choice}
        \STATE Generate perturbed set $D_{\text{pert}}$ via \S\ref{sec:oost}
        \ELSE
        \STATE Generate perturbed set $D_{\text{pert}}$ via \S\ref{sec:sgpc}
        \ENDIF
        \STATE Compute $CR$, $PCR$, $\Delta$, $\Phi$ using \S\ref{sec:am}
        \IF{multiple-choice}
        \STATE $\mathcal{C} \gets \begin{cases}
        \mathcal{C}_{\text{Minor}}, & \Delta \in (-1.6, -0.2] \\
        \mathcal{C}_{\text{Partial}}, & \Delta \in (-2.9, -1.6] \\
        \mathcal{C}_{\text{Severe}}, & \Delta \leq -2.9
        \end{cases}$
        \ELSE
        \STATE $\mathcal{C} \gets \begin{cases}
        \mathcal{C}_{\text{Minor}}, & \Delta \in (-2.4, -1.1] \\
        \mathcal{C}_{\text{Partial}}, & \Delta \in (-5.0, -2.4] \\
        \mathcal{C}_{\text{Severe}}, & \Delta \leq -5.0
        \end{cases}$
        \ENDIF
        \ENSURE $CR$, $PCR$, $\Delta$, $\Phi$, $\mathcal{C}$
    \end{algorithmic}
\end{algorithm}
% \end{minipage}
\end{center}

\section{Other Experiments}

\subsection{Semantic\,\&\,Lexical Similarity After Back-Translation}
\label{app:bt_similarity}

\paragraph{Setup.}
To quantify how much meaning and wording change during our \emph{caption perturbation} step (\S\ref{sec:sgpc}), we applied an \textbf{English$ \rightarrow $Chinese$ \rightarrow $English} back-translation to every caption in three validation splits -- {COCO-Caption}, {NoCaps}, and our {Vintage} dataset.  
For each original ($c$) and back-translated caption ($\tilde{c}$) we computed  

\begin{itemize}
  \item \textbf{SBERT} cosine similarity \cite{reimers2019sentence} as a sentence-level \emph{semantic} score, and  
  \item \textbf{BLEU-4} \cite{papineni2002bleu} as a token-overlap \emph{lexical} score.  
\end{itemize}  
We additionally report the Pearson correlation between the two metrics across captions within each dataset.

% \vspace{4pt}
\begin{table}[ht]
\centering
\setlength{\tabcolsep}{6pt}
\renewcommand{\arraystretch}{1.05}
    \resizebox{1\linewidth}{!}{
    \begin{tabular}{@{}lccc@{}}
    \toprule
    \textbf{Dataset} & \textbf{Avg.\ SBERT} $\uparrow$ & \textbf{Avg.\ BLEU} $\uparrow$ & \textbf{Correlation} $r$ \\ \midrule
    COCO Caption & 0.894 & 0.236 & 0.386 \\
    NoCaps       & 0.887 & 0.264 & 0.410 \\
    Vintage      & 0.914 & 0.441 & 0.423 \\ \bottomrule
    \end{tabular}
    }
\caption{Average semantic (SBERT) and lexical (BLEU-4) similarity between original and back-translated captions, together with their Pearson correlation ($r$).}
\label{tab:bt_similarity}
\end{table}
% \vspace{-4pt}

\paragraph{Key Observations.}
\begin{itemize}
  \item \textbf{High semantic preservation.}  All three datasets record SBERT scores close to~0.9, indicating that back-translation keeps the \emph{meaning} of captions largely intact; the \textsc{Vintage} split achieves the strongest preservation (0.914).
  \item \textbf{Substantial lexical variation.}  BLEU-4 values are comparatively low, showing that wording and surface forms differ considerably—consistent with the presence of synonym substitutions and syntactic reshuffling introduced by back-translation.
  \item \textbf{Weak yet positive coupling.}  Pearson correlations between the two metrics lie in the $0.38$-$0.42$ band, suggesting only a mild positive relationship: captions that keep more tokens also tend to retain semantics, but plenty of cases preserve meaning even with low lexical overlap.
\end{itemize}

These results justify using back-translation as a \emph{semantics-preserving yet lexically diversifying} perturbation in our contamination-detection pipeline.

\subsection{Sanity Check for the ``I don't know'' Instruction}
\label{app:sanity-idk}

\paragraph{Setup.}  
To verify that appending the uncertainty clause
``\emph{If you do not know the answer, output ``I don't know''.}''
effectively suppresses random guessing, we performed a pilot experiment
on 1\,000 randomly sampled questions from \textsc{MMStar}.  
All images were removed, so a truly vision-grounded model should either
fail or explicitly abstain.  
We evaluated the unimodal LLaMA2-7B language model under two settings:

\begin{itemize}
    \item \textbf{Deter}: deterministic decoding with the uncertainty instruction;  
    \item \textbf{Non-Deter}: deterministic decoding without the instruction.
\end{itemize}

\paragraph{Results.}  
Table~\ref{tab:idk-sanity} shows that the instruction causes the model to
respond ``I~don't~know'' 238 times and reduces apparent accuracy from
44.8\% to 25.6\% (a drop of 19.2\%).  
This confirms that nearly half of the seemingly correct answers in the
uninstructed setting are likely due to lucky guesses rather than genuine
reasoning, justifying our decision to include the clause in all main
experiments.

\begin{table}[h]
    \centering
    \resizebox{1\linewidth}{!}{
    \begin{tabular}{lcc}
        \toprule
        Setting & Accuracy (\%) & \# ``I don't know'' \\
        \midrule
        Deter (+\,instruction) & 25.6 & 238 \\
        Non‐Deter (-\,instruction) & 44.8 & 0 \\
        \bottomrule
    \end{tabular}}
    \caption{Effect of the uncertainty instruction on LLaMA2-7B.}
    \label{tab:idk-sanity}
\end{table}

``I don't know'' will therefore be treated as an explicit abstention in
the main study, ensuring reported accuracies reflect genuine
vision‐language capabilities rather than random chance.

\subsection{Prompt Consistency}
\label{app:prompt-consistency}
Our main experiments used a fixed prompt template across all evaluations. However, since different prompt wordings may influence model performance, we conducted additional experiments to assess the robustness of our methods. In particular, we tested two representative MLLMs (Claude-3.5-Sonnet and Gemini-1.5-Pro) on \textsc{ScienceQA} and \textsc{COCOCaption} under alternative prompt templates. Results show that while absolute accuracy values fluctuate slightly, the contamination-sensitive metric $\Delta$ remains nearly unchanged, confirming that our method is robust to prompt variations.

\begin{table*}[ht]
\centering
\small
\begin{tabular}{lccccc}
\hline
\textbf{MLLMs} & \textbf{Prompt Templates} & \textbf{CR} & \textbf{PCR} & \textbf{$\Delta$} & \textbf{$\Phi$} \\
\hline
Claude-3.5-Sonnet & Prompt 1 (Orig.) & \textbf{70.3} & \textbf{65.0} & \textbf{-5.3} & \textbf{15.3} \\
Claude-3.5-Sonnet & Prompt 2 & 70.4 (+0.1) & 65.0 (+0.0) & -5.4 (-0.1) & 15.5 (+0.2) \\
Claude-3.5-Sonnet & Prompt 3 & 70.4 (+0.1) & 65.0 (+0.0) & -5.4 (-0.1) & 15.5 (+0.2) \\
Gemini-1.5-Pro    & Prompt 1 (Orig.) & \textbf{68.5} & \textbf{67.9} & \textbf{-0.6} & \textbf{6.6} \\
Gemini-1.5-Pro    & Prompt 2 & 68.6 (+0.1) & 68.0 (+0.1) & -0.6 (+0.0) & 6.5 (-0.1) \\
Gemini-1.5-Pro    & Prompt 3 & 68.6 (+0.1) & 68.0 (+0.1) & -0.6 (+0.0) & 6.5 (-0.1) \\
\hline
\end{tabular}
\caption{Prompt consistency test on ScienceQA. $\Delta$ remains stable despite prompt variations.}
\label{tab:prompt_scienceqa}
\end{table*}

\begin{table*}[ht]
\centering
\small
\begin{tabular}{lccccc}
\hline
\textbf{MLLMs} & \textbf{Prompt Templates} & \textbf{CR} & \textbf{PCR} & \textbf{$\Delta$} & \textbf{$\Phi$} \\
\hline
Claude-3.5-Sonnet & Prompt 1 (Orig.) & \textbf{53.7} & \textbf{51.0} & \textbf{-2.7} & \textbf{21.8} \\
Claude-3.5-Sonnet & Prompt 2 & 53.9 (+0.2) & 51.1 (+0.1) & -2.8 (-0.1) & 22.0 (+0.2) \\
Claude-3.5-Sonnet & Prompt 3 & 53.8 (+0.1) & 51.1 (+0.1) & -2.7 (+0.0) & 21.9 (+0.1) \\
Gemini-1.5-Pro    & Prompt 1 (Orig.) & \textbf{57.5} & \textbf{55.3} & \textbf{-2.2} & \textbf{21.6} \\
Gemini-1.5-Pro    & Prompt 2 & 57.6 (+0.1) & 55.5 (+0.2) & -2.1 (+0.1) & 21.5 (-0.1) \\
Gemini-1.5-Pro    & Prompt 3 & 57.6 (+0.1) & 55.5 (+0.2) & -2.1 (+0.1) & 21.6 (+0.0) \\
\hline
\end{tabular}
\caption{Prompt consistency test on COCOCaption. Results show negligible effect on $\Delta$.}
\label{tab:prompt_coco}
\end{table*}

\paragraph{Option Order Sensitivity Test.}
The original prompt was formulated as follows:

\begin{quote}
\textbf{Prompt 1 (Original)} \\
Please answer the following multichoice question. \\
Question: \{question\} \\
Reply with answer only.
\end{quote}

We then compared two variants:

\begin{quote}
\textbf{Prompt 2} \\
Please respond to the following multiple-choice question. \\
Question: \{question\} \\
Provide only the answer. \\[6pt]
\textbf{Prompt 3} \\
Answer the multiple-choice question below. \\
Question: \{question\} \\
Reply with your answer only.
\end{quote}

\vspace{0.3em}
\noindent \textbf{Results on ScienceQA} are shown in Table~\ref{tab:prompt_scienceqa}. Across all prompt templates, $\Delta$ changes by at most $0.1$, highlighting stability.

\paragraph{Slot Guessing for Perturbed Caption.}
The original caption-prompt was:

\begin{quote}
\textbf{Prompt 1 (Original)} \\
Fill in the [MASK] of the following sentence in one word: \\
\{caption\} \\
Only reply with the word you fill in the [MASK].
\end{quote}

We added two reworded prompts:

\begin{quote}
\textbf{Prompt 2} \\
Complete the [MASK] in the sentence below with a single word: \\
\{caption\} \\
Respond only with the word you used to replace [MASK]. \\[6pt]
\textbf{Prompt 3} \\
Provide one word to fill in the [MASK] in the following sentence: \\
\{caption\} \\
Your reply should only include the word you have selected for [MASK].
\end{quote}

\vspace{0.3em}
\noindent \textbf{Results on COCOCaption} (Table~\ref{tab:prompt_coco}) again confirm robustness: CR and PCR shift slightly, but $\Delta$ stays virtually unaffected.

\noindent Overall, these experiments confirm that our detection framework is robust to natural prompt variations, further supporting its generality.

\section{Detailed Dataset Overlap Analysis}
\label{app:MLLMs Data}

It is \textbf{impractical} to quantify overlapping samples: 1) Many models do not release their complete training datasets publicly; instead, they only mention the data sources in their technical reports. 2) Even if we had access to complete training datasets, identifying specific overlapping samples using matching algorithms (such as exact match) remains challenging. This is because the original benchmarks might have undergone data augmentation before being used for model training, and multimodal benchmarks include images, both of which complicate the practical utility of matching algorithms. The feasible approach is \textbf{manually reviewing} the technical reports of these models to verify whether their training data overlaps with benchmarks, as shown in the table \ref{tab:overlap_analysis}.

\begin{table*}[ht]
\centering
\resizebox{\textwidth}{!}{
\begin{tabular}{|l|l|l|}
\hline
\textbf{MLLMs}         & \textbf{Multimodal Alignment/Pretraining Data}                                              & \textbf{Supervised Fine-Tuning Data}                           \\ \hline
\textbf{Phi-3-Vision}  & Alignment Data includes \textbf{FLD-5B}.                                                   & Not yet released                                                \\ \cline{2-3} 
                       & Open Images is one source of FLD-5B.                                                       &                                                                \\ \cline{2-3} 
                       & Open Images is also a source of \textbf{Nocaps}.                                           &                                                                \\ \cline{2-3} 
                       & Therefore, there is {\color{cellyellow} potential} overlap in Nocaps.                                  &                                                                \\ \hline
\textbf{VILA}          & {\color{cellgreen} No overlap}                                                                                 & Includes \textbf{RefCOCO, VQAv2, GQA}                           \\ \cline{2-3} 
                       &                                                                                             & MS COCO is a source of RefCOCO, VQAv2.                          \\ \cline{2-3} 
                       &                                                                                             & GQA’s source is Visual Genome Scene Graph, which also includes MS COCO. \\ \cline{2-3} 
                       &                                                                                             & \textbf{COCO Caption’s} source is MS COCO, and \textbf{NoCaps’} source includes COCO Caption. \\ \cline{2-3} 
                       &                                                                                             & Therefore, there is {\color{cellyellow} potential} overlap in COCO Caption and NoCaps. \\ \hline
\textbf{Idefics2}      & Alignment Data includes \textbf{SBU Captions}                                               & SFT Data includes \textbf{SBU Captions}: {\color{cellyellow} potential} overlap in COCO Caption and NoCaps. \\ \cline{2-3} 
                       & SBU Captions’ source includes Flickr                                                        &                                                                \\ \cline{2-3} 
                       & \textbf{COCO Caption’s} source includes MS COCO, and MS COCO’s source includes Flickr      &                                                                \\ \cline{2-3} 
                       & \textbf{NoCaps’} source includes COCO Caption                                              &                                                                \\ \cline{2-3} 
                       & Therefore, there is {\color{cellyellow} potential} overlap in COCO Caption and NoCaps.                &                                                                \\ \hline
\textbf{LLaVA-1.5}     & Alignment Data includes \textbf{SBU Captions}: COCO Caption and NoCaps with {\color{cellyellow} potential} overlap. & SFT Data includes \textbf{RefCOCO, VQAv2, GQA}: COCO Caption and NoCaps with {\color{cellyellow} potential} overlap. \\ \hline
\textbf{Yi-VL}         & Alignment Data includes \textbf{Flickr, VQAv2, RefCOCO}: & SFT Data includes \textbf{GQA}: COCO Caption and NoCaps with {\color{cellyellow} potential} overlap. \\ \cline{2-3} & COCO Caption and NoCaps with {\color{cellyellow} potential} overlap. &  \\ \hline
\textbf{DeepSeek-VL2}  & {\color{cellgreen} No overlap}                                                                                 & SFT Data includes \textbf{Flickr, GQA}: COCO Caption and NoCaps with {\color{cellyellow} potential} overlap. \\ \hline
\textbf{Qwen-VL-Chat}  & Directly uses \textbf{COCO Caption} in the pretraining stage, & Not yet released \\ \cline{2-3} & therefore there is {\color{cellred} partial or entire} overlap in COCO Caption and NoCaps. & \\ \hline
\textbf{InternVL2}     & Alignment Data includes \textbf{COCO Caption}: {\color{cellred} partial or entire} overlap in COCO Caption and NoCaps. & SFT Data includes \textbf{ScienceQA}, therefore there is {\color{cellred} partial or entire overlap} in ScienceQA. \\ \hline
\end{tabular}
}
\caption{Comparison of MLLMs and Their Data Sources}
\label{tab:overlap_analysis}
\end{table*}

\end{CJK}
\end{document}